\documentclass{article}

\usepackage{graphicx}
\graphicspath{ {./images/} }

\usepackage{arxiv}

\usepackage[utf8]{inputenc} 
\usepackage[T1]{fontenc}    
\usepackage{hyperref}       
\usepackage{url}            
\usepackage{booktabs}       
\usepackage{amsfonts}       
\usepackage{nicefrac}       
\usepackage{microtype}      
\usepackage{mathtools}      
\usepackage{dirtree}
\usepackage{listings}       
\usepackage{graphicx}
\usepackage{natbib}
\usepackage{doi}

\usepackage{listings}

\title{A modular architecture for creating multimodal agents}

\author{ \href{0000-0002-1260-3373}{\includegraphics[scale=0.06]{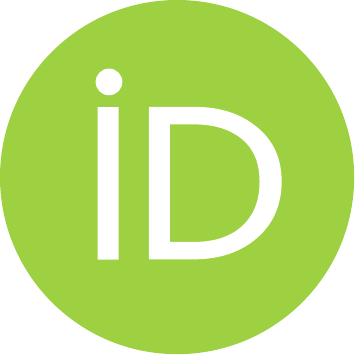}\hspace{1mm}Thomas Baier} \\
	Faculty of Humanities\\
	Vrije Universiteit Amsterdam\\
	Amsterdam, The Netherlands\\
	\texttt{t.baier@vu.nl} \\
		\And
	\href{0000-0003-3808-2614}{\includegraphics[scale=0.06]{orcid.pdf}\hspace{1mm}Selene Baez Santamaria} \\
	Faculty of Humanities\\
	Vrije Universiteit Amsterdam\\
	Amsterdam, The Netherlands\\
	\texttt{s.baezsantamaria@vu.nl} \\
	\And
	\href{0000-0002-6238-5941}{\includegraphics[scale=0.06]{orcid.pdf}\hspace{1mm}Piek Vossen} \\
	Faculty of Humanities\\
	Vrije Universiteit Amsterdam\\
	Amsterdam, The Netherlands \\
	\texttt{p.t.j.m.vossen@vu.nl} \\
}

\renewcommand{\shorttitle}{\textit{A modular architecture for creating multimodal agents}}

\hypersetup{
pdftitle={A modular architecture for creating multimodal agents},
pdfauthor={Thomas Baier, Selene Santamaria Baez, Piek Vossen},
pdfkeywords={multimodal, interactive agents, ai},
}

\begin{document}
\maketitle

\begin{abstract}
The paper describes a flexible and modular platform to create multimodal interactive agents. The platform operates through an event-bus on which signals and interpretations are posted in a sequence in time. Different sensors and interpretation components can be integrated by defining their input and output as \textit{topics}, which results in a logical workflow for further interpretations. We explain a broad range of components that have been developed so far and integrated into a range of interactive agents. We also explain how the actual interaction is recorded as multimodal data as well as in a so-called episodic Knowledge Graph. By analysing the recorded interaction, we can analyse and compare different agents and agent components.
\end{abstract}


\section{Introduction}
\vspace{-0.2cm}
Interaction amongst agents, both human and AI, is especially complex when it occurs in real-world contexts. It involves a complex psycho-social process between the agents with their intentions, capabilities, and roles, and it relates to the shared surroundings and space within which the interaction occurs. To study such interactions, test different agents, and share data and systems, two things are essential: 1) we need to be able to record the interaction as multimodal data, and 2) we need to be able to freely experiment with different components to test their impact on the interaction. This is specifically challenging because multimodal signals, such as images, sound, speech, faces and their expressions and gestures are not aligned, noisy and can be interpreted in many different ways partially dependent on each other. Building and testing systems that deal with such complex contexts requires a multitude of different technologies and software components that need to work together in a single framework to make sense.

Previously, \cite{baez2021emissor} introduced a) EMISSOR for storing sequences of multimodal signals with their annotations and b) episodic Knowledge Graphs (eKGs) for storing the accumulation of interpretations as symbolic triples. EMISSOR registers interactions as scenarios in which multimodal signals align using a temporal ruler to mark signals' start and endpoints. Segments within each signal are annotated with interpretations, such as detected objects or faces through cameras or entities and properties mentioned in speech. For example, recorded images in time can be annoted as a human face that is next identified as "Carl", who is the speaker of an audio signal (human voice), annotated as a text signal "I am from Amsterdam" which is further processed for its content. A visualization of an interaction represented in EMISSOR is included in appendix \ref{sec:emissor-visualization}.

Episodic Knowledge Graphs (eKGs) represent the accumulation of interpretations of signals over time as RDF\footnote{Resource Description Framework, \url{https://www.w3.org/RDF/}} triples consisting of subject, predicate and object identifiers (so-called Internationalized Resource Identifiers or IRIs) e.g., the speaker \textit{Carl} claiming: [leolaniWorld:Carl, n2mu:live-in, leolaniWorld:Amsterdam]. Such triples combine into more complex and rich knowledge structures and can easily be connected with other external knowledge such as DBpedia\footnote{\url{https://www.dbpedia.org}} specifying what is \textit{Amsterdam} through another triple. By interpreting conversations and perceptions as triples, the agent not only records signals but also learns about people and the world.

These components capture multimodal interactions both as the interpretation of discrete signals in a specific modality and as an integrated symbolic representation of situations in time. As explained in \citep{baez2021emissor}, EMISSOR aligns dispatched multimodal signals through a temporal ruler to which each signal is grounded. What is seen, the person Carl or any object, can be related to what is said by whom even at different points in time. Whereas EMISSOR saves temporarily grounded signals as a stream of data on disk with meta data, the "awareness of situations" and the "content of communication" are additionally pushed to the knowledge graph as perceptions, mentions, and properties, where we use JSON-LD to connect data across EMISSOR and the eKG. The knowledge in the eKG is not only represented as a collection of claims and perceptions, but also as a process, that tracks when what was claimed or perceived by whom and with which certainty, belief, or emotion. Thus, the graph represents an episodic memory\footnote{Note that the agent's memory does not forget anything and is limitless} whose time frame is aligned with the temporal ruler of EMISSOR. 

This paper describes a modular platform for creating different interactive agents whose interactions can be registered in both EMISSOR and eKGS. The platform's core is an event-bus, which can be considered a \textit{multi-topic} input/output queue that connects time-grounded signals with interpretation components. Component APIs define the expected input and and output event types, and the topics a component is connected to, such as "audio-in" or "text-out", can be defined freely and specified in the configuration of an agent. Likewise, different agents can be created easily by assembling components and configuring their connection to the event-bus.

The primary signals themselves originate from backend components that read data from sensors at time points and add them to a queue in the event-bus as certain topics. Different interpretation components consume these signal events (as defined by their input topic) and publish their output topic to the event-bus as new events. By defining a simple API to and from the event-bus, adding new signals and processing components is straightforward. These components can annotate segments from signals with interpretations, such as human speech or text, objects, and people perceived, ultimately pushing these interpretations to the eKG.

This paper explains the overall architecture and implementation of the event-bus and demonstrates how it renders interaction data to the EMISSOR format and eKGs. We also explain how different interactive agents can be created within the same platform using various components. Currently, we implemented a simple Eliza agent\footnote{\url{https://github.com/leolani/eliza-parent}} as well as a more complex Leolani agent~\citep{context19-leolani}\footnote{\url{https://github.com/leolani/leolani-mmai-parent}}. Different agents still render compatible multimodal data from their interactions within our framework. Therefore, our platform can be used to compare agents by analysing the rendered data, both in EMISSOR and as an eKG.

This paper is further structured as follows. In Section \ref{sec:event-bus}, we explain the overall architecture of the platform and its connection to EMISSOR and eKGs. Section \ref{sec:modules} describes the components that are currently available. Section \ref{sec:contributing} explains how you can create your own agent either by combining existing components or by defining your own component. All our code is available on Github under the Apache2.0 open source license from: \url{https://github.com/leolani}. 

\vspace{-0.2cm}
\section{Event-Bus architecture}
\label{sec:event-bus}
\begin{figure}[ht]
\centering
\includegraphics[scale=0.2]{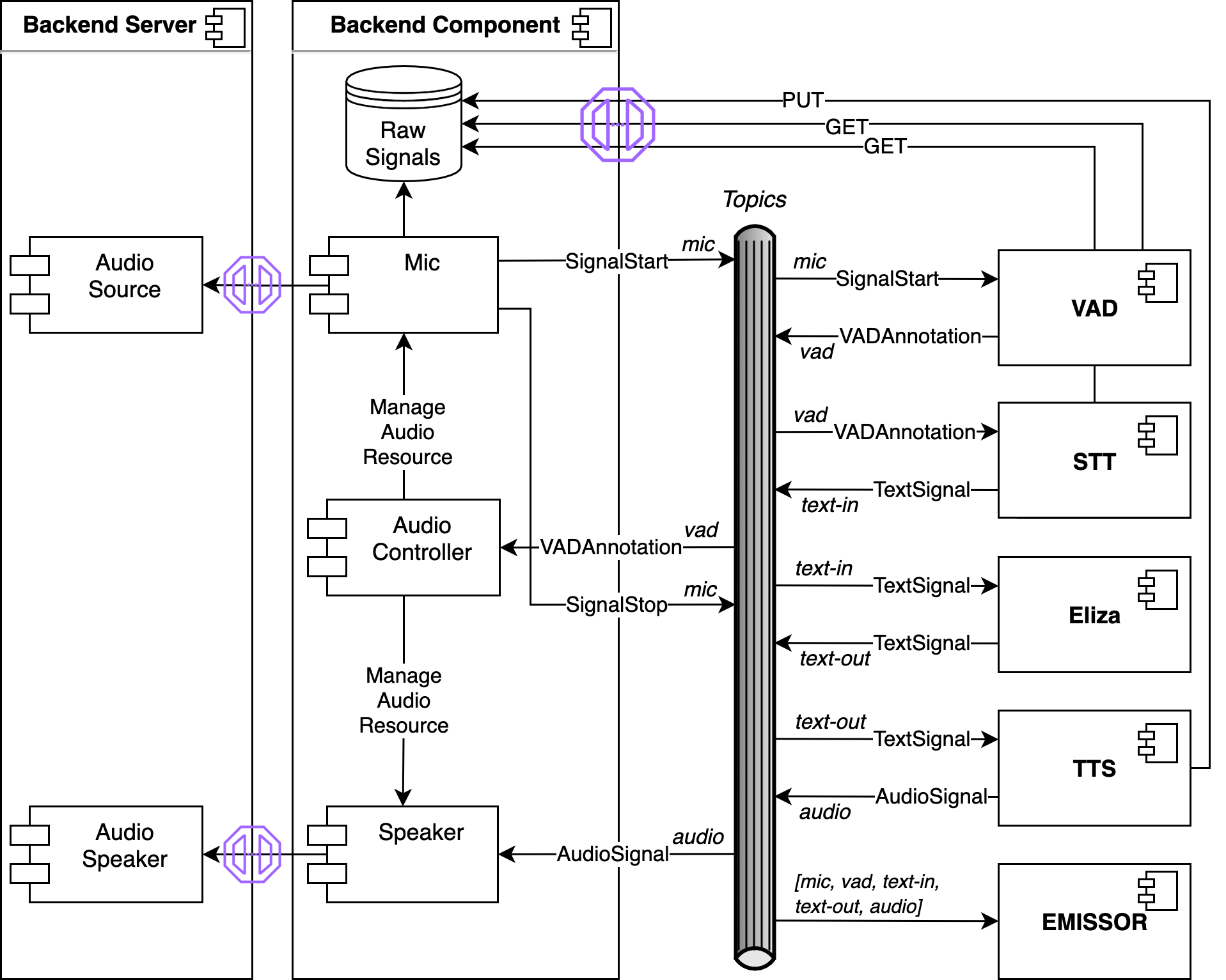}

\caption{Schematic representation of the event-bus architecture showing backend components to read from sensors and fill the queue in the event-bus and interpretation components for annotating published signals. In this schema, only audio signals are processed with an Eliza module for generating a response. Interaction data is recorded from events by the EMISSOR component.}
\label{fig:eventbus}
\end{figure}

\vspace{-0.2cm}
Figure~\ref{fig:eventbus} shows a schematic representation of the event-bus architecture. The central component is the event-bus itself, to which signals and interpretations are pushed either by backend components (at the left side) or interpretation components (at the right side). The event-bus is represented as a vertical (gray) bus to emphasize that it has a temporal dimension where events are processed in a sequence. The labels on the arrows between the event-bus and the components denote the type of event transmitted. Vertical lines in the event-bus symbolize different topics used by the individual components to define the event flow through the application. The topic labels components are connected to are denoted in italic next to the arrows.

Examples of backend components are microphones, audio controllers, and cameras. The sensors connected to backend components can be directly linked to the system or obtained from remote servers. This schema shows how a microphone produces an audio signal with a start and stop point. Some examples of interpretation components are shown on the right side of the event-bus. The VAD component at the top applies voice-activity detection to an audio signal published by the microphone. It reports back to the event-bus whether the audio is human speech, annotating the corresponding segment in the audio. The speech-to-text (STT) component applies speech recognition to any audio signal annotated as human speech and publishes the transcript as a text signal to the event-bus. The text signal is picked up by an implementation of the Eliza chatbot \citep{weizenbaum1966eliza}, which checks it for triggers and returns a response as another text signal. Finally, the text-to-speech (TTS) component picks up the text response and converts it to an audio signal. The backend speaker listens to the event-bus for any audio signals to output. An audio controller between the microphone and speaker mutes the mic when speaking and vice-versa. Any data produced by the sensors or the components as signals is stored in a backend container. 

\vspace{-0.3cm}
\subsection{Enabling reasoning}
\vspace{-0.2cm}
The main application can function with just the event-bus and some simple components, taking in signals and generating signals as a response. This behavior, however, is primarily reactive and not reflective of the information conveyed during an interaction. If we want an agent capable of reasoning over the interpretations, we first need to \textit{capture} the interaction, including metadata and signals, together with their annotations and the knowledge conveyed such that we can reason over it. For this, we included specific components to connect the event-bus data to EMISSOR and the eKG.

\begin{figure}[ht]
\centering
\includegraphics[scale=0.14]{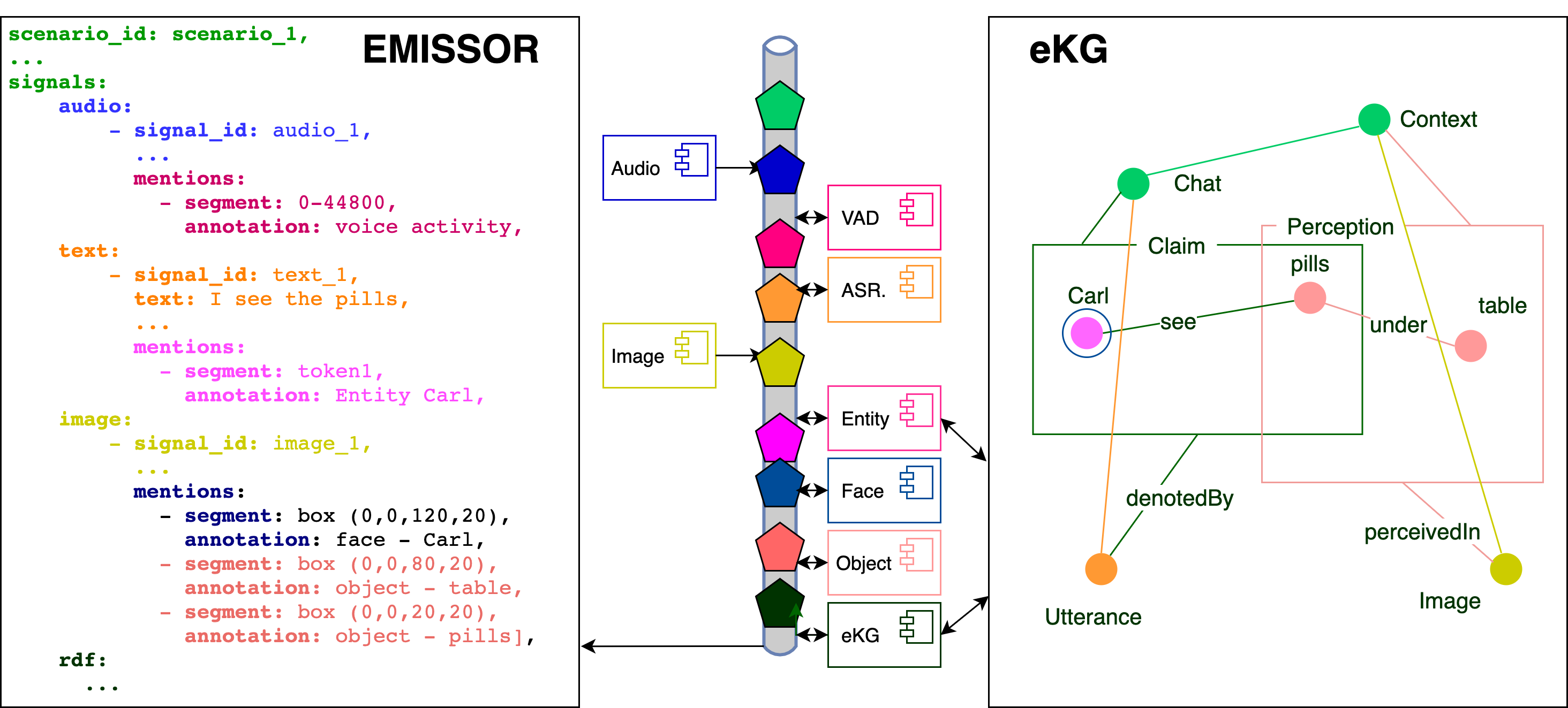}

\caption{Schematic overview of relations across data structures in EMISSOR, the episodic Knowledge Graph and the event-bus.}
\label{fig:overview}
\end{figure}

Figure \ref{fig:overview} shows the interconnections across the three components: event-bus, EMISSOR and the eKG. The three components are aligned such that each interaction in the event-bus generates a corresponding multimodal signal representation in EMISSOR and, if the interpretation yields knowledge, also an RDF triple in the eKG.

EMISSOR stores interactions as \textbf{scenarios} with metadata in JSON files for signals in each modality: images, audio, text, and RDF triples. Further details are described in \cite{baez2021emissor}. The metadata grounds the signals to a temporal ruler and defines any annotation of \textbf{segments} within a \textbf{signal} as interpretations. The signals themselves are stored as separate files on disk. We consider the RDF triples extracted from text as a modality stored on disk in EMISSOR, but they are also pushed to a separate triple store functioning as the eKG.

In the eKG model, interactions are initialized as instances of situational \texttt{eps:context}s grounded in time and space. Contexts correspond to scenarios in EMISSOR. Within a context, a series of \texttt{sem:event}s are represented, which are grounded in time like the signals in EMISSOR. We distinguish conversation and perception events. 

\paragraph{Conversation:} Conversation events in the eKG are structured as \texttt{grasp:Chat} instances containing \texttt{grasp:Utterance}s in which participants of the interaction make \texttt{grasp:Statement}s or formulate questions. The information conveyed in \texttt{grasp:Statement}s is stored in \texttt{gaf:Claim} named graphs attributed to speakers (\texttt{grasp:wasAttributedTo}) that hold certain perspectives (\texttt{grasp:Attribution}). These perspectives are properties such as \texttt{grasp:CertaintyValue}, \texttt{grasp:PolarityValue}, \texttt{grasp:SentimentValue}, \texttt{grasp:EmotionValue}.\footnote{Currently, we use NLP techniques to extract such values from the text generated by speech-recognition. In future work, modules will be added that extract these from the audio signal or face expressions as well} As such the model can express that people believe or deny certain facts with some certainty at moments in time. Claims correspond to text signals in EMISSOR, which are derived from audio signals.

\paragraph{Perceptions:} In contrast, perception events in the eKG are structured as \texttt{grasp:Visual} instances containing \texttt{grasp:Detections}s which could be objects, or people. These detections are considered to be instances of \texttt{grasp:Experience}s. Similar to conversation events, the information obtained from these \texttt{grasp:Experience}s is stored in \texttt{gaf:Claim} named graphs with triples for the output of object/face-recognition components. Therefore, the information is \texttt{grasp:wasAttributedTo} sensors, and the \texttt{grasp:Attribution}s are restricted to \texttt{grasp:CertaintyValue}s only. Perception events therefore reflect the awareness of an agent of objects and people in the context. Perceptions correspond with image signals in EMISSOR. 

As is shown in Figure \ref{fig:overview}, the event-bus supports specific components that render the data for EMISSOR and for the eKG. These components can also push data, such as annotations or triples, to the event-bus itself so that other components can take these as input for further processing. The event-bus is thus not only populated by the sensors but also by interpretation components.

In the next section, we give an overview  of all the components that are currently available in our platform and we explain in more detail how they interact. After that, we explain how new components can be added and how you can create your own agent.

\vspace{-0.2cm}
\section{Components available}
\label{sec:modules}
\vspace{-0.2cm}
The task of interaction is broken down into a large number of smaller tasks operating on specific input events taken from the event-bus and pushing their output back onto the event-bus. So far, we developed the following components, grouped by their underlying input signal:

\begin{enumerate}
    \item Audio signal
         \begin{description}
         \item[voice activity detection]\footnote{\url{https://github.com/leolani/cltl-vad}} classifying an audio signal as speech, output is an annotated audio signal pushed to the event-bus and saved to EMISSOR as an audio annotation.
         \item[automatic speech recognition]\footnote{\url{https://github.com/leolani/cltl-asr}} transforming speech to text, three different models can be chosen which generate a text output signal to the event-bus and EMISSOR.
         \item[speaker identification (under development)] identifying the speaker based on a sample of their voice, output is a speaker's identity (as an IRI and their name), and an audio signal annotated with the source pushed to the event-bus and EMISSOR but also registered in the eKG as an encounter.
     \end{description}
    \item Text signal
          \begin{description}
         \item[mention detection]\footnote{\url{https://github.com/leolani/cltl-knowledgeextraction}} detecting names and potential objects in a text signal, output is an annotated signal with named entities and object mentions, pushed to the event-bus and EMISSOR.
         \item[mention identification]\footnote{\url{https://github.com/leolani/cltl-knowledgelinking}} identifying instances in the knowledge graph for mention annotations in a text signal, output is an identifier (IRI), pushed to the event-bus and EMISSOR. Corresponding RDF triples (IRI \texttt{gaf:denotedBy} \texttt{gaf:utterance\#offset}) are pushed to the eKG.
         \item[triple extraction]\footnote{\url{https://github.com/leolani/cltl-knowledgeextraction}} detecting RDF triples in a text signal; three different modules can be selected; output are RDF triples pushed to the event-bus, EMISSOR and the eKG.
         \item[query extraction]\footnote{\url{https://github.com/leolani/cltl-questionprocessor}} detecting a question in a text signal and converting this into a SPARQL query; output is a SPARQL query send to the eKG and possibly other knowledge graphs.
         \item[emotion detection]\footnote{\url{https://github.com/leolani/cltl-emotionrecognition}} detecting basic emotions in a text signal; output is an emotion label pushed to the event-bus and EMISSOR as an annotation and possibly as a perspective to the eKG.
         \item[gesture generation]\footnote{\url{https://github.com/leolani/cltl-gesturegeneration}} returns a gesture that is an appropriate response to the emotion annotations in a text signal; output is an emotion label pushed to the event-bus and EMISSOR as an annotation and possibly as a perspective to the eKG.
         \end{description}
    \item RDF signal
       \begin{description}
         \item[knowledge representation for claims]\footnote{\url{https://github.com/leolani/cltl-knowledgerepresentation}} posting RDF claims to the eKG; output is JSON-LD as responses to the changes in the eKG by digesting new claims pushed to the event-bus.
         \item[knowledge representation for queries]\footnote{\url{https://github.com/leolani/cltl-knowledgerepresentation}} posting SPARQL queries to the eKG or other graphs; output are RDF triples as the result of the query pushed to the event-bus.
         \item[response generation]\footnote{\url{https://github.com/leolani/cltl-languagegeneration}} verbalising qualitative reflections on changes in the eKG (responses); output is a natural language response as a text signal, pushed to the event-bus and stored in EMISSOR.
         \item[language generation]\footnote{\url{https://github.com/leolani/cltl-languagegeneration}} verbalising SPARQL query results; output is natural language as a text signal, pushed to the event-bus and stored in EMISSOR.
       \end{description}
    \item Image signal
          \begin{description}
         \item[object recognition]\footnote{\url{https://github.com/leolani/cltl-object-recognition}} detecting multiple objects in images; output is bounding boxes and object labels as an annotated image signal, pushed to the event-bus and stored in EMISSOR.
         \item[face detection]\footnote{\url{https://github.com/tae898/age-gender}} detecting human faces in an image, the output are bounding boxes with the estimated age and gender as an annotated image, pushed to the event-bus and stored in EMISSOR.
         \item[face identification]\footnote{\url{https://github.com/leolani/cltl-face-recognition}} identifying people from their face, the output is an identifier (IRI and a label as name), pushed to the event-bus and stored in EMISSOR. Encounters are also registered in the eKG as perception events.
     \end{description}
\end{enumerate}

\vspace{-0.3cm}
\subsection{Processing audio signals}
\vspace{-0.2cm}
The components that process the audio signal were discussed in detail in Section 
\ref{sec:event-bus} when explaining Figure \ref{fig:eventbus}. In a nutshell, the \textbf{voice-activity-detection} component detect portions of human speech in audio, which the \textbf{automatic-speech-recognition} transcribes. Eventually, these components push a text signal to the event-bus, which is rendered from an audio signal for text processing components.

\vspace{-0.3cm}
\subsection{Processing text and rdf signals}
\vspace{-0.2cm}
The components that operate on text signals are more complex. They generate different interpretations and types of output signals and combine with components that operate on RDF triple signals. Various Natural Language Processing (NLP) modules are called within different components, among which spaCy\citep{vasiliev2020natural}, NLTK, StanfordNLP \citep{angeli2015leveraging} and various transformer models. In addition to their basic processing, \textbf{triple-extraction} components output various annotations, for example, mentions of people and objects for which referring expressions need to be resolved to people and objects known. Referring expressions can be names (Carl), common noun phrases (the waiter), or ambiguous pronouns. By reasoning over the context and previous encounters, \textbf{mention-identification} components aim to establish the referent of these expressions, possibly involving the human interlocutor in case of doubt (e.g., by posting a question as a text signal "Which waiter?"). Resolved expressions are converted to IRIs incorporated in triples. These can be registrations of mentions, e.g., \textit{Carl talking about Carla}, or as part of triples expressing a property from an individual: \texttt{leolaniWorld:Carla} \texttt{leolaniWorld:live-in} \texttt{leolaniWorld:Amsterdam}.

Whenever a \texttt{gaf:Claim} is posted to the eKG, the \textbf{knowledge-representation-for-claims} component generates a reflective response to the claim. This involves running various pre-coded SPARQL queries to detect knowledge gaps, conflicts, uncertainties, novelties, analogies, and possible generalizations. The \textbf{response-generation} component selects a result to formulate a response attempting to improve the eKG, e.g., fill gaps or resolve uncertainty. The JSON triples are verbalized as natural language text in a text signal. 

The previous route through the event-bus applies to signals classified as statements. If a text signal in the event-bus is classified as a question, the \textbf{query-extraction} component extracts a SPARQL query from the text signal that the \textbf{knowledge-representation-for-queries} component posts this to the eKG (or any other triple store). 
The \textbf{language-generation} component takes the query's result (one, many, or none triples) to verbalize it as a text signal. Different answers are generated according to the type and number of query results. Furthermore, answers consider the status of the knowledge, e.g. how certain, who is the source, and when was it mentioned.

In addition, some components only annotate mentions or emotions expressed in the text (without necessarily being involved in a triple). The \textbf{mention-detection} component detects things that (can) exist in the physical world, which are entities that the object recognition can classify. Mentions are saved as annotations of the text signal in EMISSOR and pushed to the event-bus for further processing. Next, the \textbf{mention-identification} component picks up mentions and tries to resolve its identity given the entities registered in the eKG, e.g. by checking names, the contexts, or resolving pronouns' co-reference. Identities are pushed to the event-bus and registered as \texttt{gaf:Mention}s in the eKG. Possibly, a new identity can be established and a new IRI is stored in the eKG with the properties that can be inferred.

The \textbf{emotion-detection} component interprets a text signal and pushes an emotion to the event-bus and as annotation in EMISSOR. Emotions expressed by the interlocutor represent potential perspectives of the source on the situation for the eKG. 
The same component can be applied to text signals of the agent. Especially when creating a response to new information, the corresponding emotion can be expressed by the \textbf{gesture-generation} component or audio properties such as volume and pitch.

\vspace{-0.3cm}
\subsection{Processing image signals}
\vspace{-0.2cm}
Finally, the \textbf{object recognition} and \textbf{face-detection} components take image signals as input to detect objects and faces with bounding boxes. These are saved as annotations of image signals in EMISSOR and pushed to the event-bus. From the event-bus, the \textbf{face-identification} component establishes the identity of people by comparing it with the faces of known people. Unknown faces are saved as new identities with inferred properties, such as gender and age. A new identity pushed to the event-bus is picked up by a component that asks for a person's name. The image-signal-processing components either produce annotations pushed to the event-bus and/or encoded in EMISSOR or produce an IRI with properties: name, age, gender, and registered perception event in the eKG. 

\vspace{-0.3cm}
\subsection{Sample input-output signal sequences}
\vspace{-0.2cm}
We show some schematic representations of the sequence of input and output events with their payload type and input/output topics denoted above the arrows\footnote{Topic names are exemplary and can be configured in the application}.

\begin{enumerate}
\item audio (signal)
    $\xrightarrow{\text{mic}}$ speech (annotation)
    $\xrightarrow{\text{vad}}$ text (annotation) \\
    $\xrightarrow{\text{text-in}}$ mention (annotation)
    $\xrightarrow{\text{entities}}$ IRI (annotation)
    $\xrightarrow{\text{linking}}$ claim triple (eKG)
    $\xrightarrow{\text{triples}}$ response triples (eKG)\\
    $\xrightarrow{\text{response}}$ text (signal)
    $\xrightarrow{\text{text-out}}$ audio (signal)
    \\
\item audio (signal)
    $\xrightarrow{\text{mic}}$ speech (annotation)
    $\xrightarrow{\text{vad}}$ text (annotation) \\
    $\xrightarrow{\text{text-in}}$ mention (annotation)
    $\xrightarrow{\text{entities}}$ IRI (annotation)
    $\xrightarrow{\text{linking}}$ query triple (eKG)
    $\xrightarrow{\text{triples}}$ response triples (eKG) \\
    $\xrightarrow{\text{response}}$ text (signal)
    $\xrightarrow{\text{text-out}}$ audio (signal)
    \\
\item audio (signal)
    $\xrightarrow{\text{mic}}$ speaker (annotation)
    $\xrightarrow{\text{speaker}}$ IRI (annotation)
    $\xrightarrow{\text{identity}}$ perception (eKG)
    \\
\item image (signal)
    $\xrightarrow{\text{cam}}$ face (annotation)
    $\xrightarrow{\text{face}}$ IRI (annotation)
    $\xrightarrow{\text{identity}}$ perception (eKG)
    \\
\item image (signal)
    $\xrightarrow{\text{cam}}$ object (annotation)
    $\xrightarrow{\text{object}}$ IRI (annotation)
    $\xrightarrow{\text{identity}}$ perception (eKG)
\end{enumerate}

The above components become active when the input-topic requirements are met. As such they do not represent higher-level goals or targets but are data-driven, i.e. always respond to the presence of input topics in the event-bus. The event-bus architecture however also allows to define higher-level intentions that represent behaviour of the agent to fulfill a given task or reach a goal. These intentions can be prioritised and remain active until resolved. For each intention, we can specify which components are actively listening to the event-bus and which follow-up intentions are published once a task is completed or a certain goal is reached. The above components thus are only activated within these intentions.

We implemented the following high-level intentions in the platform:

\begin{description}
    \item[Greeting] After detecting a new human face in an image signal, the agent tries to identify the person to great him/her by name or to get to know a new person by asking for a name.
    \item[Giving consent] Ask people's permission to keep the data and share it for research. If no consent is given, the agent will remove the episodic data (EMISSOR and eKG) after the interaction.
    \item[Eliza] Takes text signals as input and generate a text response using the Eliza trigger patterns and hard-coded responses.
    \item[About agent] Takes text signals as input that contain a question about the agent, and generates a text signal that informs the participant about the agent.
    \item[What do you see] Takes text signals as input that contain a question about the visual context and generates a text signal that describes what has been seen in the recent context (where recency is defined in the configuration).
    \item[Leolani] Interaction with an eKG as described in the beginning of this section, relating interpretations to identities (IRIs) and triples \citep{context19-leolani}.
    \item[Blenderbot] Process text signals with Blenderbot \citep{roller2020recipes}. Blenderbot is a generative model trained with different conversational data to generate a human-like response.
    \item[Goodbye] When a face is no longer detected or the text signal contains a goodbye clue it ends the scenario and says goodbye to the participant.
\end{description}

Both components and higher-level intentions can be created and combined freely. Partly, this provides control over certain goals or tasks to be completed but it also allows for flexible and spontaneous behavior. In the next section, we explain how you can design your own agent by adding components and high-level intentions. All the code for the platform with the currently developed components are available on Github: \url{https://github.com/leolani} under Apache2.0 license. A good starting point for building an agent is the repository: \url{https://github.com/leolani/cltl-combot}.

\vspace{-0.2cm}
\section{Developing your own agent}
\label{sec:contributing}
\vspace{-0.2cm}
As components are self-contained and loosely coupled through the event-bus, developing an agent is reduced to composing the necessary components and defining a flow of events, which is done by configuring the input and output topics of the components appropriately.

\vspace{-0.3cm}
\subsection{Adding, replacing and mixing components}
\label{sec:apps}
\vspace{-0.2cm}
To enable compatibility between components, we use EMISSOR also as the preferred data format of the payload carried by the events. Signal components publish EMISSOR {\em Signals}, eventually split up into a start and stop event. Interpretation components publish EMISSOR {\em Mentions}, defining segments and their annotations. Like this, any component that can process e.g. EMISSOR text signals or interpretations of a certain type, can be integrated into an agent by simply configuring it to listen to a topic that provides events with a text signal or the desired type of interpretation as payload. Furthermore, data from components that provide EMISSOR-based event payloads can be directly recorded in EMISSOR by a central storage component without additional adaption.

As long as component implementations process the same type of events, one implementation can simply be swapped for another. Even multiple implementations of the same component can be included in the same application, either for comparison or to increase performance. In the current version of the platform, we have for example various components for extracting triples and for speech recognition that can be applied simultaneously to increase recall or precision through voting. The source of signals and annotations is included in all EMISSOR data structures, which allows downstream components to determine the origin of an event.

Our modular architecture aims for easy integration of components that were not designed for our platform. The overhead to turn a software application into a component compatible with our platform consists of converting input and output of the application from and to EMISSOR format and connecting it to the event-bus. This can usually be achieved by adding these steps on top of existing code, without the need for modification.

\vspace{-0.3cm}
\subsection{Application types}
\label{sec:runtime}
\vspace{-0.2cm}
By choosing different implementations of the event-bus (local vs. remote), agent implementations can reach from e.g. a monolithic Python application to a containerized setup, running each component as a separate process on a single machine, for instance using {\em docker-compose}\footnote{\url{https://docs.docker.com/compose}}, or even running in a cluster consisting of many physical machines using e.g. {\em Kubernetes}\footnote{\url{https://kubernetes.io}}. These different setups do not require any code adaption of the components itself, as those only communicate through the event-bus interface, which remains identical. In a containerized setup, our architecture also allows mixing components from different platforms, e.g. using Python next to Java components, or components running different Python environments in the same agent to resolve dependency conflicts, which is a common problem in software development.

\vspace{-0.3cm}
\subsection{Out of the box}
\label{sec:available}
\vspace{-0.2cm}
For the components listed in Section \ref{sec:modules} we provide Python implementations, each in its own repository published in the Leolani Github organization \footnote{\url{https://github.com/leolani}}. Each component can be packaged as Python package and needs to provide a service module containing the EMISSOR and event-bus integration, which can be run from a Python application. Packages also provide the possibility to reuse component code across components. In the future we plan to add Docker configurations to enable running each component in a Docker container.

Our current agent implementations are realized as monolithic Python applications, which initialize and run the service modules from the each of the included component packages, and use a local implementation of the event-bus. We provide a Python implementation of the backend server which can be run on Linux/OS X/Windows platforms\footnote{\url{https://github.com/leolani/cltl-backend}}, as well as a backend server that can be run on Pepper robots\footnote{\url{https://github.com/leolani/cltl-backend-naoqi}}. Our agents can be used as a skeleton to build further agents, and in the future we will provide {\em docker–compose} and {\em Kubernetes} setups based on component Docker\footnote{\url{https://www.docker.com}} images to run the agents as fully containerized applications. In the containerized setting we will be able to run components either locally or remotely in the cloud.

To collect all code needed for an agent we created parent repositories for each agent which contain all of the agent's components as {\em git submodules}. We provide build tooling to package each of our Python components, share the packages between components and setup Python environments with all dependencies needed to run the agent application and individual components. This process can be executed centrally from the parent repository using our build setup. Following this pattern, new agents can be created also outside of our organisation, mixing components under different governance.

To build components, the \url{https://github.com/leolani/cltl-combot} provides infrastructure libraries for event-bus integration, configuration management and resource management (locking). The \texttt{cltl.combot.infra.event} module provides our event-bus interface with a local implementation as well as an implementation based on the {\em kombu}\footnote{\url{https://github.com/celery/kombu}} library supporting the {\em AMQP}\footnote{\url{https://www.amqp.org}} protocol. This allows to connect our framework to various available messaging servers like for instance {\em RabbitMQ}\footnote{\url{https://www.rabbitmq.com}}. Furthermore, \texttt{cltl.combot.infra.topic{\textunderscore}worker.TopicWorker} provides a utility class to listen to a configurable set of topics and process incoming events sequentially in a single thread, such that only the actual processing function, accepting a single event as input, needs to be implemented by the user of the library. Our own components are structured to separate functionality, using the \texttt{cltl} namespace, from the event-bus integration, using the \texttt{cltl{\textunderscore}service} namespace.

A simple example is presented in Appendix \ref{sec:component-structure}, \ref{sec:service.py} and \ref{sec:app.py}. For a full example including configuration and resource management we provide a template component at \url{https://github.com/leolani/cltl-template} with more detailed code templates, makefiles for our build setup and in the future a Dockerfile template.  

\vspace{-0.2cm}
\section{Conclusions}
\label{sec:conclusion}
\vspace{-0.2cm}
We described a platform for creating interactive agents which also captures the interaction at the signal level, the interpretation level and the integrated knowledge level. Central to our platform is an event-bus on which signals and interpretations can be published in temporal sequence as so-called topics. By defining components as services that take certain topics as input and produce other topics as output, we have the flexibility to engineer any combination of components to create agent behaviour. Our platform also supports the specification of higher-level intentions that can be carried out to fulfill a task or goal. Such intentions define which low-level components should be active to achieve this.

Multimodal interaction is extremely complex and rich. The agents that are currently built through our platform are far from what is needed to deal with all facets of human interaction. A lot of research and development is needed not only on individual components but also on their interaction and integration. Our platform will specifically help developing such complex integrated agents.

In the near future, we will use the platform to create different agents that can be used in interaction experiments. The rendered multimodal data and eKGs can be analysed and compared to assess and evaluate the agents and their components. As such, we hope it will function as a laboratory for agent development and testing.

\newpage
\bibliographystyle{unsrtnat}
\bibliography{references}

\newpage
\appendix
\section{EMISSOR multimodal data representation}
\label{sec:emissor-visualization}

\begin{figure}[ht]
\centering
\includegraphics[scale=0.14]{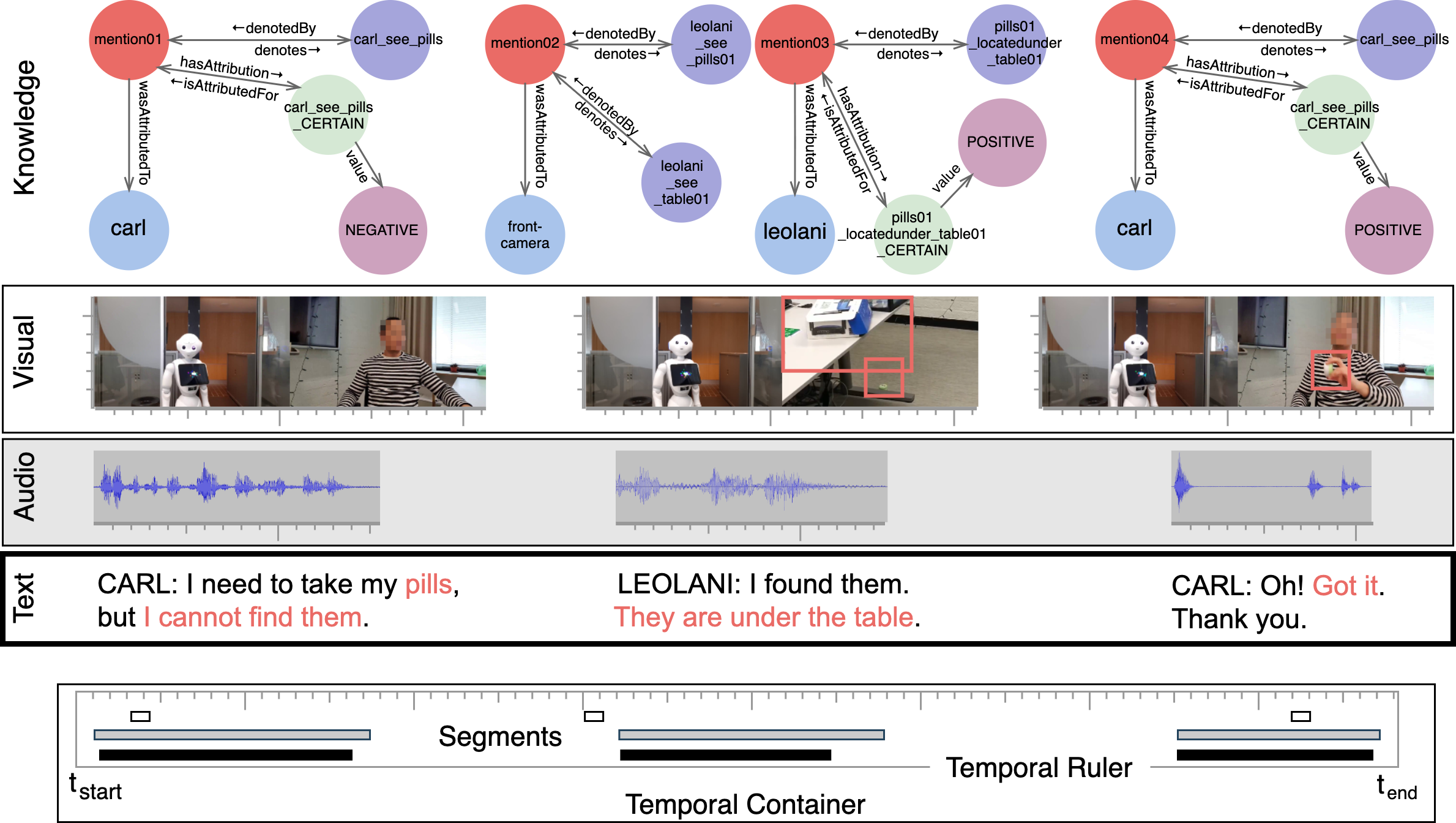}
\caption{Visualization of an interaction in the EMISSOR data format.}
\label{fig:emissor}
\end{figure}

Figure \ref{fig:emissor}, taken from \citep{baez2021emissor}, shows a visualization of four modalities (text, audio, visual, and knowledge). Signals are grounded in a temporal container on the horizontal axis, with bars marking alignments through the temporal ruler. Red boxes mark \textit{segments} annotated as \textit{mentions} of objects (pills and the table). Text \textit{segments} highlighted in red are annotated as \textit{mentions} of triples. The upper graphs represent corresponding triples from eKG generated from the annotated source modalities along the temporal sequence. The visual modality shows two different camera viewpoints (left is what Carl sees and right is what Leolani sees) concatenated side by side.

\newpage
\section{Example component structure}
\label{sec:component-structure}
\dirtree{%
.1 cltl-mycomponent.
.2 config/.
.2 requirements.txt.
.2 setup.py.
.2 src/.
.3 app.py.
.4 cltl/.
.5 mycomponent/.
.6 api.py.
.6 implementation.py.
.4 cltl{\textunderscore}service/.
.5 mycomponent/.
.6 service.py.
}

\section{Example \texttt{service.py} module}
\label{sec:service.py}

\scriptsize{
\begin{lstlisting}[language=Python,firstnumber=1]
from cltl.combot.infra.event import Event, EventBus
from cltl.combot.infra.time_util import timestamp_now
from cltl.combot.infra.topic_worker import TopicWorker
from cltl.combot.event.emissor import TextSignalEvent
from emissor.representation.scenario import TextSignal

class SimpleService:
    def __init__(self, input_topic: str, output_topic: str, event_bus: EventBus):
        self._input_topic = input_topic
        self._output_topic = output_topic
        self._event_bus = event_bus
        self._topic_worker = None

    def start(self):
        self._topic_worker = TopicWorker([self._input_topic], self._event_bus,
                                         provides=[self._output_topic],
                                         processor=self._process)
        self._topic_worker.start().wait()

    def stop(self):
        if not self._topic_worker:
            pass

        self._topic_worker.stop()
        self._topic_worker.await_stop()
        self._topic_worker = None

    def _process(self, event: Event[TextSignalEvent]):
        input = event.payload.signal.text)

        response = "" ### PROCESS THE INPUT

        if response:
            output_event = self._create_payload(response)
            self._event_bus.publish(self._output_topic, Event.for_payload(output_event))

    def _create_payload(self, response):
        signal = TextSignal.for_scenario(None, timestamp_now(), timestamp_now(), None, response)

        return TextSignalEvent.create(signal)
\end{lstlisting}
}

\section{Example \texttt{app.py} application}
\label{sec:app.py}

\scriptsize{
\begin{lstlisting}[language=Python,firstnumber=1]
from cltl.combot.infra.event.memory import SynchronousEventBus
from cltl_service.mycomponent import SimpleService

event_bus = SynchronousEventBus()
service = SimpleService("text_in", "text_out", event_bus)
try:
    service.start()
    while True:
        time.sleep(1)
except KeyboardInterrupt:
    self.service.stop()
\end{lstlisting}
}

\end{document}